\documentclass{article}

\usepackage [final] {colm2026_conference}

\usepackage{microtype}
\usepackage{graphicx}
\usepackage{booktabs}
\usepackage{amsmath}
\usepackage{amssymb}
\usepackage{xcolor}
\usepackage{hyperref}
\usepackage{url}
\usepackage{multirow}
\usepackage{enumitem}

\graphicspath{{./}}

\title{Beyond Context Windows: Persistent Discovery Context\\for Data-Centric Agents}

\author{Jalal Mahmud\\
Megagon Labs\\
Mountain View, CA, USA\\
\texttt{jalal@megagon.ai}
}

\begin{document}
\maketitle
\lhead{Published at the Context Beyond the Window Workshop at COLM 2026}

\begin{abstract}
Data-centric agents repeatedly perform a discovery step before planning
or execution: identifying the data objects relevant to a task. Yet
successful discovery outcomes are typically discarded rather than
reused. We introduce \emph{persistent discovery context}, a lightweight
memory layer that stores prior intent-to-object mappings and reuses
them to augment future retrieval. Across three structured data
environments, persistent discovery context consistently improves
retrieval quality over metadata-only search, remains effective with
automatically generated memories, and exposes a reproducible
interference failure mode. In lexically sparse domains, memory-only
retrieval can even outperform metadata-based retrieval. These findings
suggest that discovery outcomes constitute a useful form of reusable
context for data-centric agents.
\end{abstract}

\section{Introduction}
\label{sec:intro}

Agent memory systems store conversations \citep{packer2023memgpt},
observations \citep{park2023generative}, executable skills
\citep{wang2024voyager}, and reusable workflows
\citep{wang2025agentworkflow}. These mechanisms help agents retain
information across tasks and sessions. Data-centric agents face a
different challenge: before planning or execution, they must determine
which data objects are relevant to a task. Although many tasks require
similar tables and attributes, discovery is typically treated as a
per-task operation. Relevant objects are identified, used once, and
discarded rather than retained as reusable context. This raises a
simple question: \emph{Should successful discovery outcomes persist
across tasks?}

We define a discovery outcome as a mapping between a task intent and
the data objects that proved useful for accomplishing that task.
Existing metadata catalogs and schema-linking methods support
discovery \citep{datahub,yu2018spider,wang2020ratsql,
pourreza2023dinsql} but do not preserve successful discovery outcomes
across tasks.

To address this gap, we introduce \emph{persistent discovery context},
a lightweight memory layer that stores prior intent-to-object mappings
and reuses them to augment future retrieval. We evaluate the approach
on three real-world structured data environments across 125 held-out
tasks. Our findings are \textbf{fourfold}. First, persistent discovery context
consistently improves retrieval quality over metadata-only retrieval
across all three domains. Second, the approach remains effective when initialized with
LLM-generated memories, showing that useful gains can still be
obtained despite imperfect memory construction.
Third, we identify a reproducible
interference failure mode in which semantically related but incorrect
memories degrade retrieval quality. We additionally observe that in
lexically sparse domains, memory-only retrieval can outperform
metadata-based retrieval, suggesting that accumulated discovery
experience can sometimes provide a stronger signal than object
descriptions alone.

\section{Persistent Discovery Context}
\label{sec:system}
We assume a registry of data objects (tables and attributes)
augmented with natural-language metadata. Registry search retrieves candidate objects using standard text-based
retrieval over this metadata.

\textbf{Discovery Memory.} A discovery memory $m = \langle \texttt{intent},\; \texttt{objects} \rangle$ records a prior mapping between a natural-language task description and the set of data objects that proved relevant.
Discovery memories are persisted at the registry level and accumulate
across tasks within the same data environment.
Memories are created from prior discovery outcomes and reused across related tasks.

\textbf{Memory-Augmented Retrieval.} Given a query $q$, registry retrieval produces a relevance score
$r(q,o_i)$ for each candidate object $o_i$.
We then retrieve the most similar discovery memories and use them to
augment object rankings.

\begin{equation}
\text{score}(q,o_i)
=
r(q,o_i)
+
\alpha
\sum_{m \in \text{top-}K_m}
\text{sim}(q,m.\texttt{intent})
\cdot
\mathbf{1}[o_i \in m.\texttt{objects}]
\label{eq:score}
\end{equation}

where $\alpha$ controls the influence of memory and
$\mathbf{1}[o_i \in m.\texttt{objects}]$ indicates whether object
$o_i$ appeared in a retrieved memory.
Setting $\alpha=0$ recovers standard registry search.
Setting $r(q,o_i)=0$ yields a memory-only ranker (S2).

\section{Experiment}
\label{sec:exp}

\paragraph{Datasets and Tasks}
We evaluate on three real-world structured data environments from
structurally distinct domains.

\textbf{Finance.} Czech PKDD'99 banking dataset \citep{czech1999berka}:
6 tables, 42 attributes covering client accounts, transactions, and loans.

\textbf{NYC Collisions.} NYPD Motor Vehicle Collisions \citep{nyc_collisions}:
3 tables, 41 attributes covering crash-, vehicle-, and person-level records.

\textbf{Northwind.} Microsoft Northwind Traders \citep{northwind}:
8 tables, 43 attributes covering customers, orders, products,
employees, suppliers, and categories.

For each domain, we construct a registry augmented with
natural-language descriptions manually authored from schema
names, relationships, and dataset documentation. The same
descriptions are used across all retrieval conditions.
We organize tasks into 15 task families (5 per domain). Each family groups queries that share the same relevant data objects but vary in
surface vocabulary, defined thematically based on domain knowledge
(e.g., \textit{transaction\_behavior}, \textit{loan\_risk},
\textit{injury\_severity}). Each family contains one seed task used to
initialize discovery memory and a set of held-out evaluation tasks,
ranging from 5 to 12 tasks per family.
To estimate sensitivity to seed selection, we additionally perform a leave-one-out (LOO) rotation in which every task within a family serves as the seed in turn; results are reported in Appendix~\ref{sec:loo}.
Across all domains, the benchmark contains 140 tasks, comprising
125 held-out evaluation tasks and 15 seed tasks. We additionally
construct three interference tasks used only for the analysis in
Table~\ref{tab:neg}. Each task is manually annotated with 3--5 relevant data objects.

\paragraph{Experimental Conditions}

We compare four retrieval settings:

\begin{itemize}[leftmargin=*]
\item \textbf{S0: Raw Metadata.} Search over object names only.
\item \textbf{S1: Registry.} Search over object names and descriptions.
\item \textbf{S2: Memory.} Objects ranked using discovery memory
alone, without registry retrieval (i.e., $r(q,o_i)=0$ in
Equation~\ref{eq:score}).
\item \textbf{S3: Registry+Memory.} S1 augmented with discovery memory (Eq.~\ref{eq:score}).
\end{itemize}

Unless otherwise noted, S0--S3 use TF-IDF retrieval. Memory similarity uses the same retrieval model as the registry
retriever. We set $K_m{=}5$; since each domain has 5 task families
with one seed memory each, this retrieves all stored memories weighted
by similarity.
To evaluate
retriever independence, we also evaluate S1, S2, and S3 using a
neural retriever based on \texttt{all-MiniLM-L6-v2} sentence
embeddings \citep{reimers2019sbert}; these conditions are denoted
S1$^\star$, S2$^\star$, and S3$^\star$. We additionally evaluate all conditions
(S0--S3) with BM25 as an
alternative lexical retriever and compare against a naive-injection
variant (S4); both are reported in Appendix~\ref{sec:bm25} and
Appendix~\ref{sec:naive}.
To test robustness to imperfect memories, we replace the manually annotated intent-to-object mappings used to initialize discovery memory with mappings generated automatically by Claude Haiku (Appendix~\ref{sec:llmmem}).

\begin{table}[t]
  \centering
  \small
  \setlength{\tabcolsep}{5pt}
  \caption{F1@5 on held-out evaluation tasks. Unstarred: TF-IDF;
$^\star$: neural (\texttt{all-MiniLM-L6-v2}). Memory conditions use
$\alpha=0.5$. Full metrics in Appendix~\ref{sec:fullresults};
LOO robustness in Appendix~\ref{sec:loo}.}
  \label{tab:main}
  \begin{tabular}{lrrr}
    \toprule
    \textbf{Method} & \textbf{Finance} & \textbf{NYC Col.} & \textbf{Northwind} \\
    \midrule
    S0: Raw Metadata      & 0.194 & 0.227 & 0.206 \\
    S1: Registry          & 0.396 & 0.299 & 0.446 \\
    S2: Memory            & 0.358 & 0.412 & 0.386 \\
    S3: Reg.+Mem.         & 0.499 & 0.482 & 0.528 \\
    \midrule
    S1$^\star$: Registry  & 0.427 & 0.435 & 0.480 \\
    S2$^\star$: Memory    & 0.533 & 0.418 & 0.494 \\
    S3$^\star$: Reg.+Mem. & \textbf{0.594} & \textbf{0.486} & \textbf{0.599} \\
    \bottomrule
  \end{tabular}
\end{table}

\paragraph{Main Results}

Table~\ref{tab:main} summarizes the results.

\textbf{Persistent discovery context improves retrieval.}
Adding discovery memory (S3) improves F1@5 across all three domains
using TF-IDF retrieval: finance $0.396\to0.499$ ($+0.103$),
NYC collisions $0.299\to0.482$ ($+0.183$),
and Northwind $0.446\to0.528$ ($+0.082$). Appendix~\ref{sec:agent} illustrates the downstream impact on agent planning quality. The largest gain occurs in NYC collisions ($+0.183$ from S1 to S3
vs.\ $+0.072$ from S0 to S1), suggesting that memory compensates for
lexical mismatches that metadata descriptions alone cannot resolve.

With neural retrieval, memory improves F1@5 by $+0.167$ in finance,
$+0.119$ in Northwind, and $+0.051$ in NYC collisions, where the neural
base retriever is already strong (S1$^\star$ $\mathrm{MRR}=0.848$;
Appendix~\ref{sec:fullresults}). The smaller gain in NYC is consistent
with the neural retriever already resolving much of the lexical opacity
that discovery memory compensates for in the TF-IDF setting. This
pattern is confirmed by the memory-weight ablation
(Appendix~\ref{sec:alpha}), which shows that lower $\alpha$ values
perform better when the base retriever leaves less room for memory to
contribute.

\textbf{Memory alone is informative in lexically sparse domains.}
Under TF-IDF retrieval, S2 (Memory only) surpasses S1 (Registry) in
NYC collisions ($0.412$ vs.\ $0.299$), showing that the memory signal
alone outperforms metadata search when attribute names are compound and
lexically opaque (e.g., \texttt{contributing\_factor\_vehicle\_1}).
In finance and Northwind under TF-IDF, S1 remains stronger than S2,
so combining both signals (S3) is necessary to benefit from memory.
Under neural retrieval the pattern partially inverts: S2$^\star$
outperforms S1$^\star$ in finance ($0.533$ vs.\ $0.427$) and Northwind
($0.494$ vs.\ $0.480$), while S1$^\star$ remains stronger in NYC
collisions ($0.435$ vs.\ $0.418$).

Memory gains are not uniform across task families. Families with
distinctive object vocabularies (e.g., temporal analysis or transaction
behavior) benefit most from discovery memory, while families whose
objects recur across multiple task types exhibit near-zero or negative
gains (Appendix~\ref{sec:family}).

\textbf{LLM-generated memories remain useful.}
Automatically generated memories achieve object-level agreement
(F1) of 0.698, 0.762, and 0.735 (finance, NYC, Northwind) against
manually annotated discovery memories.
Retrieval still outperforms S1 in two domains ($\mathrm{F1@5}=0.433$
vs.\ $0.396$ in finance; $0.452$ vs.\ $0.299$ in NYC) and remains
above S0 in all three, indicating robustness to imperfect memory
construction (Appendix~\ref{sec:llmmem} for details).

\paragraph{Negative Result: Cross-Family Interference}
Table~\ref{tab:neg} illustrates a consistent failure mode across all
three domains. When a semantically similar but incorrect discovery
memory is activated, retrieval quality falls below the no-memory
baseline. For example, in finance, the term \emph{withdrawal} can activate a
transaction-behavior memory when the task actually concerns loan
analysis. In NYC collisions, \emph{vehicle type} can activate a
vehicle-profile memory when the task concerns injury analysis. Similar
effects occur in Northwind, where order-related vocabulary appears
across multiple task families. These results show that discovery memory is not universally beneficial.
Surface-level vocabulary overlap can activate incorrect memories and
distort rankings when the appropriate task family has not yet been
observed. This failure mode motivates future work on memory selection,
confidence estimation, and interference-aware retrieval.

\begin{table}[t]
  \centering
  \small
  \setlength{\tabcolsep}{5pt}
  \caption{Cross-family interference. Each task is evaluated under S1 (no memory),
  S3 seeded with an incorrect family memory, and S3 seeded with the
  correct family memory.}
  \label{tab:neg}
  \begin{tabular}{llrrr}
    \toprule
    \textbf{Domain} & \textbf{Interfering token(s)} & \textbf{S1} & \textbf{S3 (wrong)} & \textbf{S3 (correct)} \\
    \midrule
    Finance &
    \textit{withdrawal} $\to$ \texttt{trans.operation}
    & 0.444 & 0.222 & 0.667 \\
    NYC Collisions &
    \textit{vehicle type} $\to$ \texttt{vehicle\_type}, \texttt{vehicles}
    & 0.400 & 0.200 & 0.800 \\
    Northwind &
    \textit{order quantity} $\to$ \texttt{order\_details.quantity}
    & 0.444 & 0.000 & 0.889 \\
    \bottomrule
  \end{tabular}
\end{table}

\section{Related Work}
\label{sec:related}
Agent memory systems store reusable artifacts such as
conversational context and knowledge (MemGPT)
\citep{packer2023memgpt}, observations and reflections
(Generative Agents) \citep{park2023generative}, executable
skills (Voyager) \citep{wang2024voyager}, and workflows
(Agent Workflow Memory) \citep{wang2025agentworkflow}.
AgentSM \citep{agentsm2026} and MERIT \citep{merit2026}
extend memory to data-centric agents by storing execution
experience for schema reasoning, planning, and SQL
generation. In contrast, we study a smaller memory artifact:
discovery memory, represented as reusable intent-to-object
mappings. Rather than storing trajectories or reasoning
processes, discovery memory records which data objects proved
relevant for a task and evaluates these discovery outcomes as
a retrieval problem (R@5, MRR, F1@5) rather than through
downstream planning, SQL generation, or execution.

Our work is also related to schema linking and Text-to-SQL
\citep{yu2018spider,wang2020ratsql,pourreza2023dinsql},
knowledge-augmented SQL systems such as KAT-SQL
\citep{katsql2025}, metadata catalogs such as DataHub
\citep{datahub}, and data-discovery systems such as Starmie
\citep{starmie2023}. These systems focus on schema linking,
reasoning, query generation, metadata management, or learning
better object representations. In contrast, we investigate
whether successful discovery outcomes themselves should
persist as reusable intent-to-object mappings across tasks.

The closest conceptual connection is relevance feedback
\citep{rocchio1971,salton1990improving}, which reuses historical
relevance signals to improve future retrieval. Discovery memory differs
in that it operates on agent-generated object selections rather than
user interaction signals, transfers across semantically related but
lexically distinct tasks rather than refining a single query, and
captures reusable mappings between task intents and sets of structured
objects rather than document-level relevance.

\section{Discussion and Conclusion}
\label{sec:conclusion}

Across three structurally distinct domains, persistent discovery
context consistently improves retrieval quality over metadata-only
search across both lexical and neural retrievers and remains effective
with automatically generated memories. These findings suggest that
successful discovery outcomes constitute a reusable memory artifact
that complements conversation, experience, and workflow memory for
data-centric agents.

The primary limitation is memory interference: semantically similar
but incorrect memories can degrade retrieval below the no-memory
baseline. Future work includes interference-aware memory selection,
memory aging, larger-scale evaluation across more domains and tasks,
and evaluation using agent-generated discovery traces. More broadly,
these findings suggest that agent memory may extend beyond
conversations and workflows to include reusable knowledge about how
to navigate complex data environments.

\bibliography{references}
\bibliographystyle{colm2026_conference}

\appendix

\section{Dataset and Registry Statistics}
The finance registry contains 6 tables and 42 attributes. The NYC collisions registry contains 3 tables and 41 attributes. The Northwind registry contains 8 tables and 43 attributes.

Across three domains we construct 140 tasks in total: 125 held-out evaluation tasks organized into 15 task families (5 per domain), and 15 seed tasks (one per family) used only to populate discovery memory. We additionally construct 3 cross-family interference tasks (one per domain).

\section{Full Retrieval Results}
\label{sec:fullresults}

Table~\ref{tab:full} reports R@5, R@10, MRR, and F1@5 for all
conditions. The F1@5 column matches Table~\ref{tab:main} in the main paper.

\begin{table}[h]
  \centering
  \small
  \setlength{\tabcolsep}{4.5pt}
  \caption{Full retrieval results on held-out evaluation tasks.
Unstarred: TF-IDF; $^\star$: neural (\texttt{all-MiniLM-L6-v2}).
All memory conditions use $\alpha=0.5$.}
  \label{tab:full}
  \begin{tabular}{llrrrr}
    \toprule
    & \textbf{Method} & \textbf{R@5} & \textbf{R@10} & \textbf{MRR} & \textbf{F1@5} \\
    \midrule
    \multirow{7}{*}{\rotatebox{90}{\textit{Finance}}}
    & S0: Raw Metadata          & 0.216 & 0.377 & 0.458 & 0.194 \\
    & S1: Registry              & 0.443 & 0.618 & 0.746 & 0.396 \\
    & S2: Memory                & 0.382 & 0.623 & 0.595 & 0.358 \\
    & S3: Reg.+Mem.             & 0.551 & 0.693 & 0.816 & 0.499 \\
    \cmidrule{2-6}
    & S1$^\star$: Registry      & 0.479 & 0.675 & 0.766 & 0.427 \\
    & S2$^\star$: Memory        & 0.587 & 0.793 & 0.805 & 0.533 \\
    & S3$^\star$: Reg.+Mem.     & \textbf{0.657} & \textbf{0.846} & \textbf{0.883} & \textbf{0.594} \\
    \midrule
    \multirow{7}{*}{\rotatebox{90}{\textit{NYC Col.}}}
    & S0: Raw Metadata          & 0.267 & 0.400 & 0.348 & 0.227 \\
    & S1: Registry              & 0.340 & 0.505 & 0.648 & 0.299 \\
    & S2: Memory                & 0.476 & 0.596 & 0.749 & 0.412 \\
    & S3: Reg.+Mem.             & 0.552 & 0.751 & 0.818 & 0.482 \\
    \cmidrule{2-6}
    & S1$^\star$: Registry      & 0.501 & 0.683 & 0.848 & 0.435 \\
    & S2$^\star$: Memory        & 0.485 & 0.636 & 0.790 & 0.418 \\
    & S3$^\star$: Reg.+Mem.     & \textbf{0.559} & \textbf{0.690} & \textbf{0.799} & \textbf{0.486} \\
    \midrule
    \multirow{7}{*}{\rotatebox{90}{\textit{Northwind}}}
    & S0: Raw Metadata          & 0.234 & 0.327 & 0.604 & 0.206 \\
    & S1: Registry              & 0.501 & 0.682 & 0.795 & 0.446 \\
    & S2: Memory                & 0.421 & 0.530 & 0.612 & 0.386 \\
    & S3: Reg.+Mem.             & 0.586 & 0.792 & 0.886 & 0.528 \\
    \cmidrule{2-6}
    & S1$^\star$: Registry      & 0.534 & 0.744 & 0.792 & 0.480 \\
    & S2$^\star$: Memory        & 0.539 & 0.726 & 0.653 & 0.494 \\
    & S3$^\star$: Reg.+Mem.     & \textbf{0.661} & \textbf{0.827} & \textbf{0.806} & \textbf{0.599} \\
    \bottomrule
  \end{tabular}
\end{table}

\section{LLM-Generated Discovery Memories}
\label{sec:llmmem}

To evaluate robustness to imperfect memories, we replace manually
annotated discovery memories with memories generated automatically by
Claude Haiku. For each seed task, the model receives the task
description and registry contents and is asked to identify the
relevant data objects.

Generated memories are evaluated against manually annotated discovery
memories using object-level F1. We then use the generated memories to
initialize discovery memory and measure downstream retrieval F1@5:

\begin{itemize}
  \item Finance: memory agreement $\mathrm{F1}=0.698$; S3 $\mathrm{F1@5}=0.433$ vs.\ S1 $\mathrm{F1@5}=0.396$
  \item NYC Collisions: memory agreement $\mathrm{F1}=0.762$; S3 $\mathrm{F1@5}=0.452$ vs.\ S1 $\mathrm{F1@5}=0.299$
  \item Northwind: memory agreement $\mathrm{F1}=0.735$; S3 $\mathrm{F1@5}=0.435$ vs.\ S1 $\mathrm{F1@5}=0.446$
\end{itemize}

Finance and NYC show clear improvement over registry-only search.
Northwind LLM memories do not improve over S1, reflecting noisier
object selections in a domain with higher cross-entity vocabulary
overlap (e.g., order-related terms appearing across multiple task
families).

Across all three domains, retrieval quality remains above the
raw-metadata baseline (S0), indicating robustness to imperfect memory
construction.

\section{Per-Family Analysis}
\label{sec:family}

Memory gains are not uniform across task families.

\begin{figure}[h]
\centering
\includegraphics[width=0.95\linewidth]{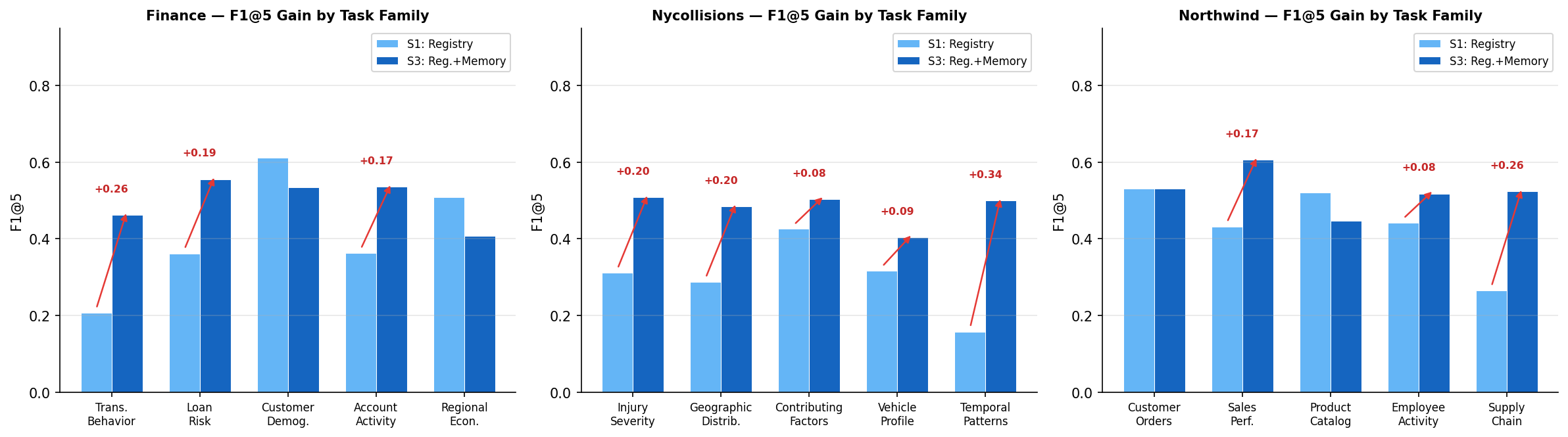}
\caption{F1@5 improvement (S1$\rightarrow$S3) by task family.}
\label{fig:family_appendix}
\end{figure}

Families with distinctive object vocabularies benefit most from
discovery memory. The top three families---temporal analysis
($+0.34$), supply chain ($+0.26$), and transaction behavior
($+0.26$)---show the largest gains, while families such as loan risk
($+0.19$) and account activity ($+0.17$) show moderate improvement.

In contrast, families whose objects appear across multiple task types
show near-zero or negative gains. Customer demographics ($-0.077$),
regional economics ($-0.102$), and product catalog ($-0.074$) all
degrade slightly relative to registry-only retrieval. In these cases,
the seed memory introduces objects that overlap with unrelated queries,
adding a bias that the registry score cannot fully overcome.

These results suggest that discovery memory is most effective when the
relationship between task intent and relevant objects is specific and
stable, and least effective---or mildly harmful---when multiple task
families share similar object sets.

\section{Memory Weight Ablation}
\label{sec:alpha}

\begin{figure}[h]
\centering
\includegraphics[width=0.95\linewidth]{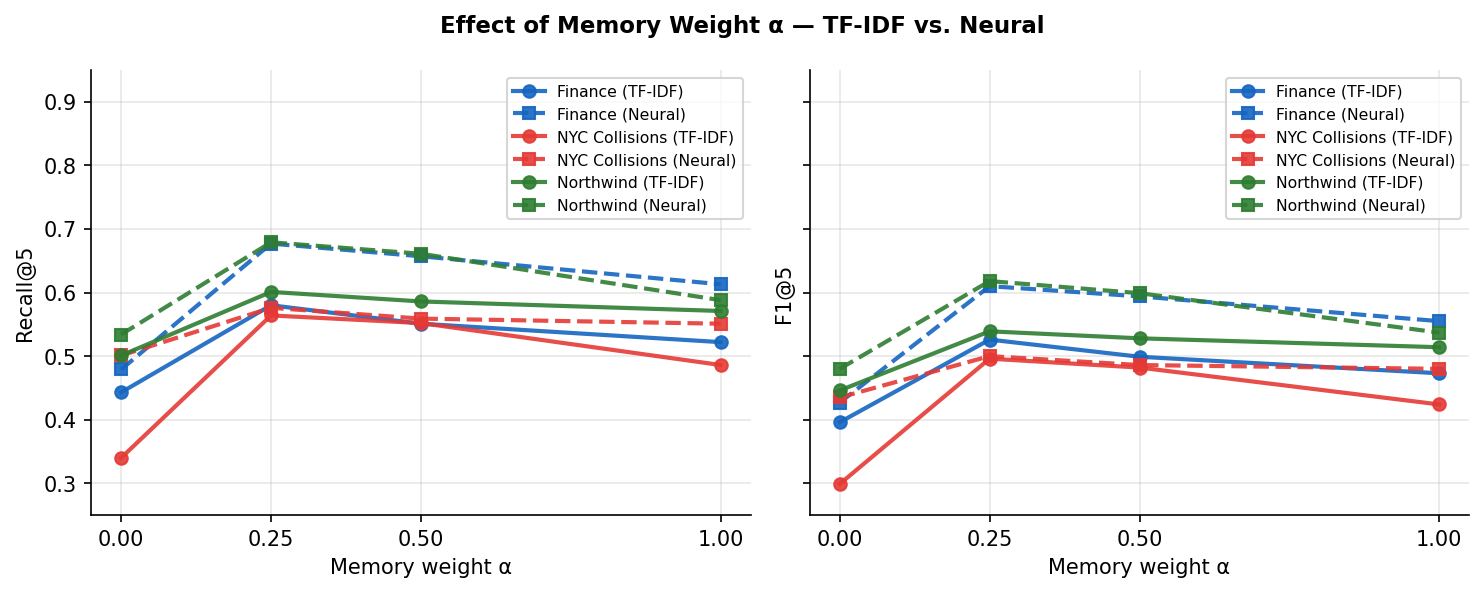}
\caption{Recall@5 and F1@5 as a function of memory weight $\alpha$
across three domains and two retrievers. $\alpha{=}0$ recovers
registry-only (S1); $\alpha{=}1$ gives memory full control. Both
retrievers peak at $\alpha{=}0.25$; over-weighting memory degrades
performance in all settings.}
\label{fig:alpha}
\end{figure}

As shown in Figure~\ref{fig:alpha}, both TF-IDF and neural retrieval
peak at $\alpha=0.25$ across all three domains. The absolute gain from
the $\alpha=0$ baseline to the peak is substantially smaller for neural
retrieval (e.g., NYC collisions: $+0.197$ under TF-IDF vs.\ $+0.065$
under neural), consistent with a stronger base retriever leaving less
room for memory to contribute. Over-weighting memory at $\alpha=1.0$
degrades performance in all settings. Setting $\alpha=0$ recovers the
registry-only baseline. Table~1 reports results at the default
$\alpha{=}0.5$; using the ablation-optimal $\alpha{=}0.25$ yields
modest additional gains but the main conclusions are unchanged.

\section{Blending vs.\ Direct Memory Injection}
\label{sec:naive}

Table~\ref{tab:naive} compares registry+memory blending (S3) against a
direct injection baseline (S4) that promotes all objects associated
with the highest-scoring memory to the front of the ranking without
combining memory and registry scores.

\begin{table}[h]
\centering
\small
\setlength{\tabcolsep}{4.5pt}
\caption{Comparison of S3 (Registry+Memory, blended) and S4 (Naive
Injection) on held-out evaluation tasks. S3 outperforms S4 in both
MRR and F1@5 across all three domains.}
\label{tab:naive}
\begin{tabular}{llrrrr}
\toprule
\textbf{Domain} & \textbf{Method} & \textbf{R@5} & \textbf{R@10} & \textbf{MRR} & \textbf{F1@5} \\
\midrule
\multirow{2}{*}{Finance}
  & S3: Reg.+Mem.  & 0.551 & 0.693 & \textbf{0.816} & \textbf{0.499} \\
  & S4: Naive Inj. & 0.455 & 0.742 & 0.682 & 0.427 \\
\midrule
\multirow{2}{*}{NYC}
  & S3: Reg.+Mem.  & 0.552 & 0.751 & \textbf{0.818} & \textbf{0.482} \\
  & S4: Naive Inj. & 0.506 & 0.703 & 0.761 & 0.440 \\
\midrule
\multirow{2}{*}{Northwind}
  & S3: Reg.+Mem.  & 0.586 & 0.792 & \textbf{0.886} & \textbf{0.528} \\
  & S4: Naive Inj. & 0.423 & 0.749 & 0.690 & 0.387 \\
\bottomrule
\end{tabular}
\end{table}

S3 outperforms S4 in both MRR and F1@5 across all three domains.
Directly injecting objects from a retrieved memory can elevate
irrelevant objects when the memory only partially matches the query.
By combining registry and memory signals, S3 preserves retrieval
precision while still promoting objects associated with relevant prior
discovery outcomes.

\section{Leave-One-Out Seed Robustness}
\label{sec:loo}

Table~\ref{tab:loo} reports F1@5 mean and standard deviation under
leave-one-out (LOO) seed rotation. For each task family, every task
serves as the seed in turn while the remaining tasks are evaluated,
yielding 56, 51, and 36 evaluation trials for Finance, NYC
Collisions, and Northwind, respectively. The relative ordering of
methods is consistent with the fixed-seed results in
Table~\ref{tab:main}, indicating that the observed gains are not
sensitive to the choice of seed task.

\begin{table}[h]
  \centering
  \small
  \setlength{\tabcolsep}{4.5pt}
  \caption{F1@5 mean$\pm$std under leave-one-out seed rotation.
  Unstarred: TF-IDF; $^\star$: neural retriever.}
  \label{tab:loo}
  \begin{tabular}{lrrr}
    \toprule
    \textbf{Method} & \textbf{Finance} & \textbf{NYC Col.} & \textbf{Northwind} \\
    \midrule
    S0: Raw Metadata     & $0.192 \pm 0.129$ & $0.229 \pm 0.134$ & $0.188 \pm 0.087$ \\
    S1: Registry         & $0.409 \pm 0.150$ & $0.302 \pm 0.089$ & $0.482 \pm 0.092$ \\
    S2: Memory           & $0.404 \pm 0.143$ & $0.448 \pm 0.121$ & $0.337 \pm 0.132$ \\
    S3: Reg.+Mem.        & $0.539 \pm 0.094$ & $0.521 \pm 0.074$ & $0.532 \pm 0.101$ \\
    \cmidrule{1-4}
    S1$^\star$: Registry & $0.442 \pm 0.132$ & $0.458 \pm 0.081$ & $0.497 \pm 0.088$ \\
    S2$^\star$: Memory   & $0.536 \pm 0.138$ & $0.519 \pm 0.106$ & $0.458 \pm 0.129$ \\
    S3$^\star$: Reg.+Mem.& $\mathbf{0.607 \pm 0.091}$ & $\mathbf{0.614 \pm 0.072}$ & $\mathbf{0.605 \pm 0.092}$ \\
    \bottomrule
  \end{tabular}
\end{table}

\section{BM25 Retriever Comparison}
\label{sec:bm25}

Table~\ref{tab:bm25} reports S0--S3 results using Okapi BM25 in place
of TF-IDF. BM25 underperforms TF-IDF at the S1 and S3 levels across
all three domains; S0 differences are negligible. A notable gap appears at the S1 registry level
(e.g., Finance: 0.358 vs.\ 0.396; NYC: 0.292 vs.\ 0.299;
Northwind: 0.388 vs.\ 0.446), reflecting that BM25's term-frequency
saturation is poorly matched to the long, semantically rich prose
descriptions in the registry. S2 (memory-only) results are comparable
between BM25 and TF-IDF, as memory retrieval matches queries against
short intent strings rather than registry descriptions. Despite the
weaker base, S3 still improves over S1 with BM25 in all three domains,
confirming that memory boosting is beneficial regardless of the
underlying retriever.

\begin{table}[h]
  \centering
  \small
  \setlength{\tabcolsep}{4.5pt}
  \caption{BM25 retriever results on held-out evaluation tasks (S0--S3).
  TF-IDF F1@5 is shown in parentheses for comparison.}
  \label{tab:bm25}
  \begin{tabular}{llrrrr}
    \toprule
    & \textbf{Method} & \textbf{R@5} & \textbf{R@10} & \textbf{MRR} & \textbf{F1@5 (TF-IDF)} \\
    \midrule
    \multirow{4}{*}{\rotatebox{90}{\textit{Finance}}}
    & S0: Raw Metadata   & 0.198 & 0.363 & 0.414 & 0.177 \quad (0.194) \\
    & S1: Registry       & 0.395 & 0.553 & 0.668 & 0.358 \quad (0.396) \\
    & S2: Memory         & 0.418 & 0.591 & 0.607 & 0.379 \quad (0.358) \\
    & S3: Reg.+Mem.      & 0.434 & 0.584 & 0.677 & 0.393 \quad (0.499) \\
    \midrule
    \multirow{4}{*}{\rotatebox{90}{\textit{NYC Col.}}}
    & S0: Raw Metadata   & 0.260 & 0.400 & 0.333 & 0.221 \quad (0.227) \\
    & S1: Registry       & 0.335 & 0.473 & 0.594 & 0.292 \quad (0.299) \\
    & S2: Memory         & 0.464 & 0.610 & 0.752 & 0.402 \quad (0.412) \\
    & S3: Reg.+Mem.      & 0.453 & 0.661 & 0.642 & 0.392 \quad (0.482) \\
    \midrule
    \multirow{4}{*}{\rotatebox{90}{\textit{Northwind}}}
    & S0: Raw Metadata   & 0.234 & 0.327 & 0.604 & 0.206 \quad (0.206) \\
    & S1: Registry       & 0.435 & 0.630 & 0.773 & 0.388 \quad (0.446) \\
    & S2: Memory         & 0.421 & 0.533 & 0.629 & 0.386 \quad (0.386) \\
    & S3: Reg.+Mem.      & 0.477 & 0.702 & 0.790 & 0.429 \quad (0.528) \\
    \bottomrule
  \end{tabular}
\end{table}

\section{Agent-in-the-Loop Demonstration}
\label{sec:agent}

To assess downstream impact, we provide the top-5 retrieved objects
from S1 and S3 to a planning agent (Claude Haiku) and ask it to
generate an analysis plan for the task ``Find customers with unusual
cash withdrawals.''

\begin{figure}[h]
\centering
\small
\begin{tabular}{p{0.46\linewidth}|p{0.46\linewidth}}
\textbf{S1 --- Registry only} &
\textbf{S3 --- Registry+Memory} \\
\hline
\textit{Retrieved:} client\_id, birth\_date, bank, disp\_id, client &
\textit{Retrieved:} operation, \textbf{trans}, balance, type, account \\
\hline
\multicolumn{2}{p{0.96\linewidth}}{\textit{Task: ``Find customers with unusual cash withdrawals''}} \\
\hline
\textit{``Query the bank database transactions table
\textbf{(implied in the banking database)}, filtering for
cash withdrawal operations.''} &
\textit{``Query the \textbf{trans} entity where
\textbf{operation} = `cash withdrawal' and \textbf{type} = `debit'
to isolate withdrawal transactions.''} \\
\end{tabular}
\caption{Illustrative effect of retrieval context on downstream agent
planning. With S1 retrieval, the agent must infer that a transactions
table exists and works around its absence. With S3 retrieval, the agent
directly references \texttt{trans.operation} and \texttt{trans.type},
producing a more executable analysis plan.
Same model, same task; only the retrieved context changes.}
\label{fig:agent}
\end{figure}

S3 retrieval provides the exact transaction-related objects needed for
the task, while S1 retrieval contains mostly demographic objects. As a
result, the S3 plan directly references the correct operational fields
and produces a more executable analysis strategy.

\end{document}